\pdfoutput=1

\documentclass[11pt]{article}

\usepackage{acl}

\newcommand{\comb}{GandR\xspace}
\newcommand{\inputbase}{input TF-IDF\xspace}
\newcommand{\outputbase}{output TF-IDF\xspace}

\newcommand{\callintent}{\textsc{\textcolor{blue}{in:create\_call}}\xspace}
\newcommand{\groupslot}{\textsc{\textcolor{teal}{sl:group}}\xspace}
\newcommand{\contactslot}{\textsc{\textcolor{magenta}{sl:contact}}\xspace}
\newcommand{\sendmessageintent}{\textsc{\textcolor{darkgray}{in:send\_message}}\xspace}
\newcommand{\other}[1]{\textsc{\textcolor{darkgray}{#1}}\xspace}
\newcommand{\weatherintent}{\textsc{\textcolor{blue}{in:get\_weather}}\xspace}

\newcommand{\weatherarreslot}{\textsc{\textcolor{teal}{sl:weather\_attrubute}}\xspace}
\newcommand{\ordinal}{\textsc{\textcolor{magenta}{sl:ordinal}}\xspace}
\newcommand{\myself}{\textit{myself}\xspace}

\definecolor{targetmention1}{rgb}{0, 0.46484375, 0.73046875}
\definecolor{targetmention2}{rgb}{0.9296875, 0.46484375, 0.19921875}
\definecolor{targetmention3}{rgb}{0.2980, 0.7529, 0.7490}

\usepackage{times}
\usepackage{latexsym}

\usepackage[T1]{fontenc}

\usepackage[utf8]{inputenc}

\usepackage{microtype}

\usepackage{amsmath}
\usepackage{xspace}
\usepackage{booktabs}
\usepackage{url}
\usepackage{siunitx}
\usepackage{graphicx}
\usepackage{multirow}
\usepackage{tikz}
\usepackage{pgfplots}

\title{Generate-and-Retrieve: use your predictions to improve retrieval for semantic parsing}

\newcommand*\samethanks[1][\value{footnote}]{\footnotemark[#1]}

\author{
Yury Zemlyanskiy\thanks{\enskip Work primarily done at Google Research.} \ \footnotemark[2] \quad 
Michiel de Jong\samethanks[1] \ \footnotemark[2] \quad 
Joshua Ainslie\footnotemark[3] \quad 
Panupong Pasupat\footnotemark[3] \quad \\ 
{\bf Peter Shaw}\footnotemark[3] \quad
{\bf Linlu Qiu}\footnotemark[3] \quad
{\bf Sumit Sanghai}\footnotemark[3] \quad
{\bf Fei Sha}\footnotemark[3] \\[.5em]
\footnotemark[2] \footnotetext[2]{ } University of Southern California \quad \footnotemark[3] \footnotetext[3]{ }  Google Research \\
\small{\texttt{\{yury.zemlyanskiy,msdejong\}@usc.edu}} \\
\small{\texttt{\{jainslie,ppasupat,petershaw,linluqiu,sumitsanghai,fsha\}@google.com}}
}

\begin{document}
\maketitle
\begin{abstract}
A common recent approach to semantic parsing augments sequence-to-sequence models by retrieving and appending a set of training samples, called exemplars. The effectiveness of this recipe is limited by the ability to retrieve informative exemplars that help produce the correct parse, which is especially challenging in low-resource settings. Existing retrieval is commonly based on similarity of query and exemplar inputs. We propose \comb, a retrieval procedure that retrieves exemplars for which outputs are also similar. \comb first generates a preliminary prediction with input-based retrieval. Then, it retrieves exemplars with outputs similar to the preliminary prediction which are used to generate a final prediction. \comb sets the state of the art on multiple low-resource semantic parsing tasks. 
\end{abstract}

\begin{figure*}[t!]
    \centering
    \includegraphics[width=1.0\textwidth]{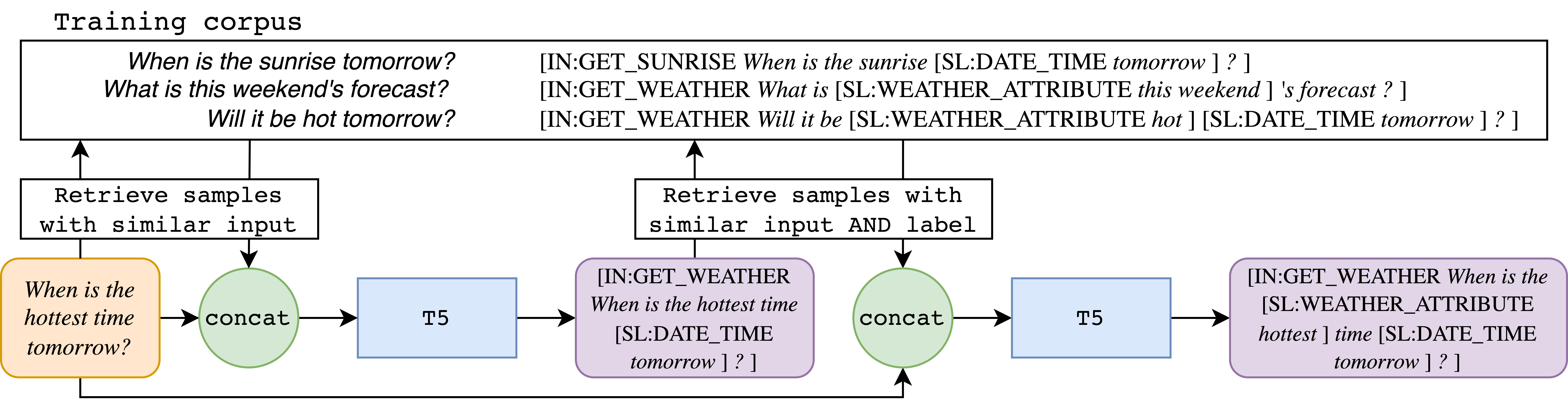}    
    \caption{Overview of \comb. First, \comb generates a preliminary prediction using an input augmented with exemplars with similar inputs. Then, \comb retrieves exemplars based on a relevance measure balancing input similarity and similarity between the preliminary prediction and exemplar outputs, and generates a final prediction based on these exemplars.}
    \label{fig:model_overview}
\end{figure*}

\section{Introduction}

A common and successful approach to structured prediction problems \citep{mtop, topv2} is to treat the gold structure as a sequence and fine-tune a sequence-to-sequence model such as T5 \citep{t5} or BART \citep{bart}. However, the performance of fine-tuned models suffers in low resource scenarios where available training data is limited relative to the complexity of the task \citep{topv2}.

Existing work \citep{casper,retronlu,reina} has found that retrieving related training samples, denoted \textit{exemplars}, and appending the retrieved input-output pairs to the sample input before processing the sample can improve performance in low resource settings. In principle, all information from exemplars is available to the model during training and could be stored in model parameters. However, in practice the model may not successfully retain all information, and reminding the model of salient input-output patterns at test time appears to help. 

That raises the question: what exemplars are most informative for the model? Existing work focuses on retrieving exemplars for which the input is partially similar to the test input, effectively answering ``\textit{What is the output for similar inputs?}''. In this work we explore whether there is complementary information in exemplars that answer the inverse question, ``\textit{What is the input for similar outputs?}''.

We propose Generate-and-Retrieve (\comb), a method to retrieve exemplars with similar output as well as input. As the true output of a sample is in general unknown, \comb proceeds in two steps. First, a preliminary prediction is generated using retrievals with similar input only. Then, a new set of exemplars is retrieved based on a relevance measure that balances the similarity of the inputs and the similarity of the preliminary prediction and the exemplar output. Figure \ref{fig:model_overview} provides an overview of the method.

We evaluate \comb in the setting of task-oriented semantic parsing, a core component of widely used virtual assistants. We show that similarity in output space provides a complementary signal to input similarity, yielding retrievals that prove more informative for the model. Moreover, for many structured prediction tasks the output space is more structured than the free-form input text, so that simple, non-learned distance measures work well for outputs even when inputs are lexically dissimilar. Table \ref{table:paper_example} demonstrates an example where our proposed similarity function retrieves an example that is somewhat less similarly phrased but with more similar output, and the model produces a better prediction as a result. Finally, the model has the opportunity to verify that its preliminary predictions are valid outputs in the target language.

The proposed method strongly improves performance in low-resource settings for semantic parsing, achieving state of the art results for low-resource and transfer benchmarks in MTOP \cite{mtop} and TopV2 \cite{topv2}.  

\begin{table*}[t!]
\centering
\begin{tabular}{>{\raggedright\arraybackslash}p{139pt} >{\centering\arraybackslash}p{47pt} >{\centering\arraybackslash}p{38pt} >{\centering\arraybackslash}p{44pt} | >{\centering\arraybackslash}p{40pt} >{\centering\arraybackslash}p{40pt}}
    \textbf{Model} & \textbf{MTOP}\textsubscript{\text{boot}} & \textbf{MTOP}\textsubscript{1k} & \textbf{MTOP}\textsubscript{25\%} & \textbf{TOPv2}\textsubscript{W} & \textbf{TOPv2}\textsubscript{R}\\
    \hline
    Reptile \citep{topv2} & & & & 77.7 & 70.5\\
    RAF \cite{retrieve_and_refill} & & & & 78.7 & \\
    CASPER \citep{casper} & 73.3 / 83.9 & & & & \\ 
    \hline
    T5 & 72.9 / 83.3 & 62.8 & 78.5 & 79.2 & 68.8\\
    T5 with input TF-IDF & 74.9 / 84.5 & 67.2 & 79.4 & 79.9 & 71.0\\
    \comb & \textbf{76.4 / 84.6} & \textbf{67.8} & \textbf{80.1} & \textbf{80.5} & \textbf{71.7}\\    
\end{tabular}
\caption{Results on semantic parsing benchmarks. We report the percentage exact match between true and predicted labels as sequences. Results are on test set for all benchmarks except MTOP\textsubscript{boot}, where we report on dev to remain comparable with CASPER.}

\label{table:top_main_results}
\end{table*}

\section{Method}

We approach semantic parsing as a conditional language generation task and apply a T5 sequence-to-sequence model \citep{t5} to predict a parse $y$ given a query $x$. For each sample, we retrieve $K=4$ relevant training exemplars sampled according to a relevance scoring function. We append the retrieved input-output pairs to the sample input and apply the T5 model to the augmented input to predict a parse output. In particular, let $(x'_1, y'_1), \ldots, (x'_K, y'_K)$ denote the retrieved input-output pairs, then the augmented input is 
\begin{align*}
    x' = x\texttt{ || }x'_1\texttt{ \& }y'_1\texttt{ || }x'_2\texttt{ \& }y'_2\texttt{ || }\ldots
\end{align*}
Our approach closely follows that of \citet{casper}, differing primarily in the choice of relevance function. During evaluation we retrieve the top $K$ most relevant exemplars. During training, we sample retrievals according to a geometric distribution over the relevance score rank. In particular, the probability that we retrieve an exemplar is given by $p (1-p)^r$ where $r$ is the rank of the exemplar according to relevance score and $p$ is a temperature hyperparameter.

In \citet{casper}, the relevance score is given by the inner product of Universal Sentence Encoder \citep{use} encodings of the candidate input and the sample input. We found that a simple TF-IDF \citep{tfidf} similarity baseline achieves comparable or better results.

Our proposed approach, \comb, builds on the input-similarity baseline by constructing a hybrid similarity measure that takes into account not only the similarity between sample and candidate inputs, but also the similarity between the sequence predicted by the model and the candidate output. See Figure \ref{fig:model_overview} for an overview. First, \comb generates a preliminary prediction using an input augmented with exemplars with similar inputs. Then, \comb retrieves exemplars based on a hybrid similarity measure over inputs and outputs, and generates a final prediction based on these exemplars.

Specifically, let $\hat{y}_i$ be preliminary prediction, then the proposed output similarity between samples $i$ and $j$ is given by the TF-IDF similarity between the predicted structure (in our case, the set of intents and slots) and the structure of the true parse $y_j$. Our proposed relevance score is a weighted sum of input and output similarity (see Appendix~\ref{appendix:method} for detailed description on how output similarity is computed): 
\begin{align*}
    R_{ij} = (1 - \alpha) \text{TF-IDF}(x_i, x_j) + \alpha \text{TF-IDF}(\hat{y}_i, y_j)
\end{align*}
\subsection{Training}
For simplicity, we train \comb in two stages. We start training with TF-IDF input relevance scoring, yielding model $M_1$. Model $M_1$ is used to generate \comb preliminary predictions during training and evaluation. We continue training $M_1$ for the remaining training steps, yielding $M_2$, which is used to generate final predictions augmented with retrievals from $M_1$. Note that this two-stage training is for convenience only, and it is possible to use a single set of weights $M_{single}$ to generate preliminary and final \comb predictions. In that case, $M_{single}$ needs to be trained with a mix of input-only and \comb retrieval augmentations to ensure it is able to use either effectively.

\section{Related Work}

Sequence-to-sequence models \citep{t5,bart} have achieved state-of-the-art performance on task-oriented semantic parsing \citep{mtop, topv2, sbtop} as well as other structured prediction tasks~\citep{t5}. The general approach is to pre-train on language modeling and perform fine-tuning on the specific domain of interest. 

Several works augment the input with retrieved exemplars from the training data, with differing methods for selecting informative examples. \citet{casper} and \citet{retronlu} retrieve exemplars with similar input encodings from a pre-trained neural encoder, evaluating on semantic parsing. \citet{reina} retrieves examplars for which the input has high BM25 similarity with the sample input, with good performance on language generation. We adopt a similar approach with TF-IDF similarity as a baseline for semantic parsing.

\citet{gpt_neo} and \citet{cbr} learn dense retrievers in the spirit of \citet{dpr}, providing another path to incorporate label information for retrieval. \citet{atlas} proposes other methods to fine-tune a dense retriever. These approaches require training a separate model specifically for retrieval, possibly with additional learning signal. In contrast, we employ a sparse similarity measure over model predictions that are produced incidentally in the course of fine-tuning the main model.

Selecting relevant training exemplars is also important for in-context prompting \cite{ppp}. Similar to related fine-tuning literature, work in this direction uses either a pre-trained \cite{promptpretrain} or fine-tuned \cite{promptfine} sentence encoder to retrieve exemplars.

\section{Experiments}

\begin{table}[ht!]
\centering
\begin{tabular}{lrrrrrr}
    \textbf{Dataset} & \textbf{\#Train} & \textbf{\#Dev} & \textbf{\#Test}\\
    \hline
    MTOP & 15667 & 2234 & 4385 \\
    MTOP\textsubscript{1k} & 1096 & 2234 & 4385 \\
    MTOP\textsubscript{25\%} & 3916 & 2234 & 4385 \\
    TOPv2\textsubscript{S} & 83703 & 11967 & 27336 \\
    TOPv2\textsubscript{W} & 176 & 147 & 5682 \\
    TOPv2\textsubscript{R} & 493 & 337 & 5767 \\
    \hline
\end{tabular}
\caption{Dataset statistics: the number of examples per dataset and split.}
\label{table:dataset_stats}
\end{table}
\begin{table}[t!]
\centering
\begin{tabular}{>{\raggedright\arraybackslash}p{90pt} >{\centering\arraybackslash}p{42pt} >{\centering\arraybackslash}p{42pt}}
    \textbf{Model} & \textbf{MTOP} & \textbf{TOPv2}\textsubscript{S} \\
    \hline
    RAF  & & 87.1\\
    CASPER & 86.4 & \\ 
    \hline
    T5& 85.7 & 86.9 \\
    T5 input TF-IDF & 86.4 & 87.0 \\
    \comb & 86.4 & 87.0\\    
\end{tabular}
\caption{Performance on high-resource settings.}
\label{table:resource_ablation}
\end{table}

\begin{figure}[h]
\centering
\begin{tikzpicture}[scale=0.75]
\begin{axis}[
xlabel={Output similarity weight, $\alpha$.},
ylabel={Exact match, \%},
mark=x,
legend pos=south east,
ymajorgrids=true,
xmajorgrids=true,
grid style=dashed
]
\addplot[mark=triangle*, mark size=3pt] table {
0    74.908005
0.25   74.842854
0.5  75.655946
0.75   76.384501
0.9  75.702682
1.0   75.465460
};
\node at (axis cs:-0.1,75.0) [anchor= south west] {input TF-IDF};
\node at (axis cs:1.1,75.4) [anchor= north east] {output TF-IDF};
\end{axis}
\end{tikzpicture}
\caption{Performance on MTOP\textsubscript{\text{boot}} development set as a function of output similarity weight $\alpha$. \label{fig:mtop_boot_vs_alpha}}
\end{figure}
\begin{figure}[h]
\centering
\begin{tikzpicture}[scale=0.75]
\begin{axis}[
xlabel={Number of retrieved exemplars, $K$.},
ylabel={Exact match, \%},
mark=x,
legend pos=south east,
ymajorgrids=true,
xmajorgrids=true,
grid style=dashed
]
\addplot[color=targetmention2,mark=square*, mark size=2pt] table {
1   65.921
2   65.9806
4   66.8904
};
\addlegendentry{input TF-IDF}
\addplot[color=targetmention1,mark=triangle*, mark size=3pt] table {
1   66.428
2   66.7263
4   67.5615
};
\addlegendentry{GandR}
\end{axis}
\end{tikzpicture}
\caption{Performance on the development set of \textbf{MTOP}\textsubscript{1k} as a function of the number of retrieved exemplars    $K$. \label{fig:mtop_1k_vs_k}}
\end{figure}
\begin{table}[t]
\centering
\begin{tabular}{>{\raggedright\arraybackslash}p{64pt}| >{\centering\arraybackslash}p{39pt} >{\centering\arraybackslash}p{31pt} >{\centering\arraybackslash}p{31pt}}
    \textbf{Retriever} & \textbf{MTOP}\textsubscript{\text{boot}} & \textbf{TOPv2}\textsubscript{W} & \textbf{TOPv2}\textsubscript{R}\\
    \hline
    input TF-IDF & 35.9 & 55.1 & 20.1 \\
    output TF-IDF & 70.3 & 74.8 & 53.7 \\
    \comb & 70.0 & 68.7 & 52.5 \\    
\end{tabular}
\caption{Template recall@K=4 on the development sets for MTOP\textsubscript{boot}, TOPv2\textsubscript{W} and TOPv2\textsubscript{R}.}
\label{table:template_recall}
\end{table}

\subsection{Setup}
We evaluate \comb and baselines on semantic parsing benchmarks MTOP \cite{mtop} and TOPv2 \cite{topv2}, focusing on low-resource and transfer settings. MTOP is a medium-sized semantic parsing dataset used in \citet{casper}, for which we evaluate on the \textit{domain bootstrapping} setting in which one of the domains is limited to a very small amount of training data. We also evaluate on low-resource settings MTOP\textsubscript{1k} and MTOP\textsubscript{25\%} in which we randomly sample 1k and 25\% of training samples, respectively. TOPv2 is centered on transfer to low-resource domains: models are trained on a set of high resource-domains denoted as TOPv2\textsubscript{S} and then fine-tuned on low-resource \textit{Weather} and \textit{Reminder} domains\footnote{We are using 25 SPIS low resource split from \citet{topv2}.}, denoted as TOPV2\textsubscript{W} and TOPV2\textsubscript{R}. We show the sizes of datasets and splits in Table~\ref{table:dataset_stats}. See Appendix~\ref{appendix:training} for details on the training setup.

\subsection{Main results}
The results of our primary experiments are shown in Table~\ref{table:top_main_results}. We find that \inputbase is a strong baseline, rivaling or improving over prior work. Further, \comb retrieval outperforms all baselines, setting the state of the art on evaluated settings.

\subsection{Ablations and discussion}

\paragraph{Retrieval is less important for high-resource settings} Table~\ref{table:resource_ablation} shows results on the high-resource full MTOP and TOPv2 datasets. In higher-resource settings, augmenting the input with exemplars appears to be both less effective and less sensitive to retrieval method, with almost identical results among methods with and without retrieval for the highest resource TOPv2 dataset.

\paragraph{Using hybrid similarity leads to better retrieval quality}
Figure~\ref{fig:mtop_boot_vs_alpha} displays MTOP\textsubscript{boot} performance as a function of TF-IDF output weight $\alpha$. The results demonstrate that input and output similarity signals are strongly complementary. See Figure~\ref{fig:mtop_1k_vs_alpha} and Figure~\ref{fig:mtop_25_vs_alpha} in the Appendix for similar experiments on the MTOP\textsubscript{1k} and MTOP\textsubscript{25\%} benchmarks.

\paragraph{Considering output similarity leads to higher template recall}
Following \citet{casper}, we compute \textit{template recall@K} as a proxy metric for retrieval. This measure corresponds to the proportion of evaluation samples for which at least one of the top K retrievals has the same template (identical intents and slots) as the gold parse. Results show (Table~\ref{table:template_recall}) that considering output as well as input similarity increases template recall. We note that \outputbase has similar or higher template recall than \comb even though it has lower performance. Ultimately, template recall is only a proxy, and we are really interested in retrieval \textit{informativeness}; \comb's performance shows that balancing input similarity and template recall leads to exemplars that are most helpful for the model.

\paragraph{Hybrid similarity helps across different numbers of retrieved exemplars} 
Figure~\ref{fig:mtop_1k_vs_k} shows that \comb outperforms \inputbase similarity on MTOP\textsubscript{1k} when we retrieve different number of exemplars: 1, 2 or 4. See Figure~\ref{fig:mtop_25_vs_k} in the Appendix for a similar experiment on the  MTOP\textsubscript{25\%} benchmark.

\subsection{Error analysis}
\begin{table}[t!]
\centering
\begin{tabular}{>{\raggedright\arraybackslash}p{0.93\columnwidth}}
\begin{center} \underline{Input sample} \end{center}
$x$: \textit{Could you connect me to the Musicals group}\\
$y$: [\callintent [\groupslot \textit{Musicals}] ]\\
\vspace{-8pt}
\begin{center} \underline{Training sample with similar input} \end{center}
$x_1$: \textit{musicals in windham this weekend} \\
$y_1$: [\other{in:get\_event} [\other{sl:category\_event} \textit{musicals} ] [\other{sl:location} \textit{windham} ] [\other{sl:date\_time} \textit{this weekend} ] ]\\
$\hat{y}$: [\callintent [\contactslot \textit{me}] [\groupslot \textit{Musicals}] ]\\
\vspace{-8pt}
\begin{center} \underline{Training sample with similar input and label} \end{center}
$x_1$: \textit{can you please send text to the development group} \\
$y_1$: [\sendmessageintent [\groupslot \textit{development}]]\\
$\hat{y}$: [\callintent [\groupslot \textit{Musicals}] ]\\
\end{tabular}
\caption{\label{table:paper_example} Input TF-IDF retrieves an exemplar with lexical overlap (`musicals') that is not relevant to the sample. The \comb retrieval balances lexical and label similarity and leads to a correct prediction. Single representative exemplar out of 4 displayed for each method. See Table~\ref{table:paper_example_full} in the Appendix for all retrieved exemplars.}
\end{table}

The primary motivation for \comb is that hybrid similarity leads to more informative exemplars. Informativeness can only be objectively measured through model performance, but our motivating intuition appears to be borne out by samples in the data. We observe a number of different cases for which output or hybrid-similarity retrieval can help. Table \ref{table:paper_example} shows an example of a case for which \inputbase retrieves an irrelevant example with lexical overlap, while \comb retrieves an example with both lexical and parse overlap, leading to a correct prediction. Using preliminary predictions for retrieval can also allow the model to verify whether its predictions are correct. A common simple case when this can help is if the model generates a prediction that is dissimilar to any samples in the training set in which case the model may reconsider whether that prediction is correct (Table \ref{table:verify_example}). Considering output similarity does come with tradeoffs. Table \ref{table:negative_example} demonstrates a situation where output similarity distracts the model away from a lexically similar and informative exemplar and the model is wrong as a result.

\section{Conclusion}

We propose \comb, a new method for structured prediction that generates a preliminary prediction, retrieves training exemplars with similar outputs (and similar inputs), and augments the input with the retrieved exemplars to generate a final prediction. We demonstrate that using output similarity yields improvements for semantic parsing in low-resource settings, achieving state of the art results on several semantic parsing benchmarks.

\section*{Acknowledgments}
We thank William Cohen, Nicholas Fitzgerald and Luke Vilnis for insightful discussions and reviewers for their feedback.
This work is partially supported by NSF Awards IIS-1513966/ 1632803/1833137, CCF-1139148, DARPA Awards\#: FA8750-18-2-0117, FA8750-19-1-0504,  DARPA-D3M - Award UCB-00009528, Google Research Awards, gifts from Facebook and Netflix, and ARO\# W911NF-12-1-0241 and W911NF-15-1-0484.


\newpage
\appendix
\section{Method}

\subsection*{Output TF-IDF}
\label{appendix:method}

We measure the similarity between two parses, denoted as output similarity, by their TF-IDF similarity. More precisely, we process parses as a sentence where each intent and slot in the parse is treated as a single token. Then we simply compute TF-IDF similarity between the constructed sentences. Slot values are discarded - the similarity between slot values should already be captured by input TF-IDF similarity.
\section{Training}
\label{appendix:training}

We initialize our model from the public T5.1.1-base checkpoint\footnote{\url{https://github.com/google-research/text-to-text-transfer-transformer/blob/main/released_checkpoints.md\#t511}}. All models are fine-tuned using the ADAM optimizer with dropout 0.1 and weight decay 0.01.  In our experiments, we consider sampling temperature $p \in \{0.5, 0.1, 0.05\}$, batch size $\in \{128, 256\}$ and output similarity weight $\alpha \in \{0, 0.25, 0.5, 0.75, 0.9, 1.0\}$, where $\alpha=0$ is our input TF-IDF baseline. 

For all experiments we train (or in case of early stopping, provide the opportunity to train) \inputbase and \comb for the same number of steps. We train for 120k steps on MTOP\textsubscript{boot} (90k first stage/30k second stage), 40k (10k/30k) on MTOP\textsubscript{1k} and 60k (30k/30k) on MTOP\textsubscript{25\%} with early stopping. For TOPv2, we train for 90k steps on the source domain followed by 7.5k (5k/2.5k) steps on TOPv2 transfer domains without early stopping.

Hyperparameters were selected based on development set performance. We measure and report average performance over 3 random seeds.  

\section{Detailed experimental results}
\begin{figure}[h]
\centering
\begin{tikzpicture}[scale=0.75]
\begin{axis}[
xlabel={Output similarity weight, $\alpha$.},
ylabel={Exact match, \%},
mark=x,
legend pos=south east,
ymajorgrids=true,
xmajorgrids=true,
grid style=dashed
]
\addplot[mark=triangle*, mark size=3pt] table {
0    66.8904
0.25   67.5615
0.5  67.4869
0.75   67.0694
0.9  67.2036
1.0   67.0246
};
\node at (axis cs:0.05,66.85) [anchor= south west] {input TF-IDF};
\node at (axis cs:1,66.99) [anchor= north east] {output TF-IDF};
\end{axis}
\end{tikzpicture}
\caption{Performance on MTOP\textsubscript{\text{1k}} development set as a function of output similarity weight $\alpha$. \label{fig:mtop_1k_vs_alpha}}
\end{figure}
\begin{figure}[h]
\centering
\begin{tikzpicture}[scale=0.75]
\begin{axis}[
xlabel={Output similarity weight, $\alpha$.},
ylabel={Exact match, \%},
mark=x,
legend pos=south east,
ymajorgrids=true,
xmajorgrids=true,
grid style=dashed
]
\addplot[mark=triangle*, mark size=3pt] table {
0    78.2849
0.25   78.8069
0.5  78.8814
0.75   78.8665
0.9  78.8665
1.0   78.6130
};
\node at (axis cs:0.05,78.25) [anchor= south west] {input TF-IDF};
\node at (axis cs:1.05,78.57) [anchor= north east] {output TF-IDF};
\end{axis}
\end{tikzpicture}
\caption{Performance on MTOP\textsubscript{\text{25}} development set as a function of output similarity weight $\alpha$. \label{fig:mtop_25_vs_alpha}}
\end{figure}
\begin{figure}[h]
\centering
\begin{tikzpicture}[scale=0.75]
\begin{axis}[
xlabel={Number of retrieved exemplars, $K$.},
ylabel={Exact match, \%},
mark=x,
legend pos=south east,
ymajorgrids=true,
xmajorgrids=true,
grid style=dashed
]
\addplot[color=targetmention2,mark=square*, mark size=2pt] table {
1   77.3304
2   77.5391
4   78.2849
};
\addlegendentry{input TF-IDF}
\addplot[color=targetmention1,mark=triangle*, mark size=3pt] table {
1   77.8524
2   78.3893
4   78.8814
};
\addlegendentry{GandR}
\end{axis}
\end{tikzpicture}
\caption{Performance on the development set of \textbf{MTOP}\textsubscript{25\%} as a function of the number of retrieved exemplars, $K$. \label{fig:mtop_25_vs_k}}
\end{figure}
\begin{table*}[h]
\centering
\begin{tabular}{lrr}
    \textbf{Model} & \textbf{New domain} & \textbf{Other domains}\\
    \hline
    \textbf{Average} \\
    CASPER & 73.3 & 83.9 \\
    T5 & 72.88 & 83.34 \\
    T5 with input TF-IDF & 74.91 & 84.50 \\
    T5 with output TF-IDF & 75.47 & 84.62 \\
    \comb & 76.38 & 84.50 \\
    \hline
    \textbf{alarm} \\
    CASPER & 77.96 & 84.53 \\
    T5 & 72.58 & 83.50 \\
    T5 with input TF-IDF & 76.88 & 84.53 \\
    T5 with output TF-IDF & 76.34 & 84.82 \\
    \comb & 76.88 & 84.77 \\
    \hline
    \textbf{calling} \\
    CASPER & 74.47 & 82.16 \\
    T5 & 77.81 & 82.63 \\
    T5 with input TF-IDF & 78.42 & 83.53 \\
    T5 with output TF-IDF & 79.64 & 83.32 \\
    \comb & 78.42 & 83.58 \\
    \hline
    \textbf{event} \\
    CASPER & 83.74 & 83.71 \\
    T5 & 85.37 & 83.14 \\
    T5 with input TF-IDF & 83.74 & 84.47 \\
    T5 with output TF-IDF & 85.37 & 84.90 \\
    \comb & 86.18 & 84.47 \\
    \hline
    \textbf{messaging} \\
    CASPER & 77.27 & 83.29 \\
    T5 & 78.41 & 82.71 \\
    T5 with input TF-IDF & 79.00 & 84.12 \\
    T5 with output TF-IDF & 79.00 & 84.17 \\
    \comb & 82.95 & 83.92 \\
    \hline
    \textbf{music} \\
    CASPER & 53.14 & 85.70 \\
    T5 & 50.24 & 84.71 \\
    T5 with input TF-IDF & 56.52 & 85.90 \\
    T5 with output TF-IDF & 57.00 & 85.90 \\
    \comb & 57.49 & 85.75 \\
    \hline
\end{tabular}
\caption{Detailed domain bootstrapping results for MTOP. \comb is for output similarity weight $\alpha=0.75$. CASPER results are from \citet{casper}.}
\label{table:mtop_domain_boot}
\end{table*}

\section{Error analysis}

We provide two additional retrieval examples (Table \ref{table:verify_example} and Table \ref{table:negative_example}) to illustrate cases where \comb retrieval can be beneficial or harmful.

\begin{table}[ht!]
\centering
\begin{tabular}{>{\raggedright\arraybackslash}p{0.93\columnwidth}}
\begin{center} \underline{Input sample} \end{center}
$x$: \textit{When will it rain next?}\\
$y$:  [\weatherintent \textit{When will it} [\weatherarreslot \textit{rain} ] \textit{next ?} ]\\
\vspace{-8pt}
\begin{center} \underline{Training sample with similar input} \end{center}
$x_1$: \textit{Will it rain all day tomorrow in New York?} \\
$y_1$: [\weatherintent \textit{Will it} [\weatherarreslot \textit{rain} ] [\other{sl:date\_time} \textit{all day tomorrow} ] in [\other{sl:location} \textit{New York} ] \textit{?} ] \\
$x_2$: \textit{When will the sun rise today?} \\
$y_2$: [\other{in:get\_sunrise} \textit{When will the sun rise} [\other{sl:date\_time} \textit{today} ] \textit{?} ]\\
$x_3$: \textit{When will it snow this year} \\
$y_3$: [\other{in:unsupported\_weather} \textit{When will it} [\weatherarreslot = \textit{snow} ] [\other{sl:date\_time} \textit{this year} ] ]\\
$x_4$: \textit{Will it be cold today?} \\
$y_4$:  [\weatherintent \textit{Will it be} [\weatherarreslot \textit{cold} ] [\other{sl:date\_time} \textit{today ?} ] ]\\
$\hat{y}$:  [\weatherintent \textit{When will it} [\weatherarreslot \textit{rain} ] [\ordinal \textit{next}] \textit{?} ]\\
\vspace{-8pt}
\begin{center} \underline{Training sample with similar input and label} \end{center}
$x_1$: \textit{Is it raining?} \\
$y_1$:  [\weatherintent \textit{Is it} [\weatherarreslot \textit{raining} ] \textit{?} ]\\
$x_2$: \textit{How cold is it outside?} \\
$y_2$: [\weatherintent \textit{How} [\weatherarreslot \textit{cold} ] \textit{is it outside ?} ]\\
$x_3$: \textit{what time will the storms roll in} \\
$y_3$: [\weatherintent \textit{what time will the} [\weatherarreslot \textit{storms} ] \textit{roll in} ]\\
$x_4$: \textit{How much snow?} \\
$y_4$: [\weatherintent \textit{How much} [\weatherarreslot \textit{snow} ] \textit{?} ]\\
$\hat{y}$:  [\weatherintent \textit{When will it} [\weatherarreslot \textit{rain} ] \textit{next ?} ]\\
\end{tabular}
\caption{\label{table:verify_example} Input TF-IDF predicts the \ordinal slot, which exists in pre-training domains but does not apply to the Weather domain. \comb has high slot coverage, so if a slot exists, it will likely be present in at least one of the retrieved exemplars. The fact that \comb does not retrieve an exemplar with the \ordinal slot (as it is not present in the Weather training exemplars) provides a hint for the model that it may be an invalid slot and \comb's updated prediction eliminates it.}
\end{table}
\begin{table}[ht!]
\centering
\begin{tabular}{>{\raggedright\arraybackslash}p{0.93\columnwidth}}
\begin{center} \underline{Input sample} \end{center}
$x$: \textit{call myself}\\
$y$:  [\callintent [\contactslot \textit{myself}]]\\

\vspace{-8pt}
\begin{center} \underline{Training sample with similar input} \end{center}
$x_1$: \textit{Things to do by myself} \\
$y_1$: [\other{in:get\_event} ] \\
$x_2$: \textit{remind myself to get breakfast} \\
$y_2$:  [\other{in:create\_reminder}  [\other{sl:person\_reminded} \textit{myself} ] [\other{sl:todo} \textit{get breakfast}]]\\
$x_3$: \textit{Reminder to make dentist app for myself for next week} \\
$y_3$: [\other{in:create\_reminder} [\other{sl:todo} \textit{make dentist app for myself} ] [\other{sl:date\_time} \textit{for next week} ] ]\\
$x_4$: \textit{place myself as being unavailable} \\
$y_4$: [\other{in:set\_unavailable} ]\\
$\hat{y}$:  [\callintent [\contactslot \textit{myself}]]\\

\vspace{-8pt}
\begin{center} \underline{Training sample with similar input and label} \end{center}

$x_1$: \textit{Call Jeremy} \\
$y_1$: [\callintent [\contactslot \textit{Jeremy}]] \\
$x_2$: \textit{Call Tammy} \\
$y_2$: [\callintent [\contactslot \textit{Tammy} ] ] \\
$x_3$: \textit{Please call Jane} \\
$y_3$: [\callintent [\contactslot \textit{Jane} ] ]
$x_4$: \textit{Make a call to Steve}
$y_4$: [\callintent [\contactslot \textit{Steve} ] ]
$\hat{y}$:  [\callintent \textit{myself} ]\\
\end{tabular}
\caption{\label{table:negative_example} Normally, the \contactslot slot for the \callintent intent is paired with a name. In this case, however, the contact is \myself. The model with input TF-IDF retrieval generates the correct slot as it retrieves another instance with slot \myself due to lexical similarity in the input. In contrast, \comb retrieves exemplars with perfectly matching templates, but without the same slot, such that it does not assign \myself to the \contactslot slot in its prediction.}
\end{table}
\begin{table}[t!]
\centering
\begin{tabular}{>{\raggedright\arraybackslash}p{0.93\columnwidth}}
\begin{center} \underline{Input sample} \end{center}
$x$: \textit{Could you connect me to the Musicals group}\\
$y$: [\callintent [\groupslot \textit{Musicals}] ]\\
\vspace{-8pt}
\begin{center} \underline{Training sample with similar input} \end{center}
$x_1$: \textit{musicals in windham this weekend} \\
$y_1$: [\other{in:get\_event} [\other{sl:category\_event} \textit{musicals} ] [\other{sl:location} \textit{windham} ] [\other{sl:date\_time} \textit{this weekend} ] ]\\
$x_2$: \textit{could you message a video} \\
$y_2$: [\sendmessageintent [\other{sl:type\_content} \textit{video} ] ]\\
$x_3$: \textit{Please could you remind me to walk the dog} \\
$y_3$: [\other{in:create\_reminder} [\other{sl:person\_reminded} \textit{me} ] [\other{sl:todo} \textit{walk the dog} ] \\
$x_4$: \textit{Could you tell me the weather in Paris?} \\
$y_4$: [\other{in:get\_weather} [\other{sl:location} \textit{Paris} ] ]
$\hat{y}$: [\callintent [\contactslot \textit{me}] [\groupslot \textit{Musicals}] ]\\
\vspace{-8pt}
\begin{center} \underline{Training sample with similar input and label} \end{center}
$x_1$: \textit{Start a call Spoken Word} \\
$y_1$: [\callintent [\groupslot \textit{Spoken Word} ] ] \\
$x_2$: \textit{can you please send message to the anime group} \\
$y_2$: [\sendmessageintent [\groupslot \textit{anime} ] ] \\
$x_3$: \textit{can you please send text to the development group} \\
$y_3$: [\sendmessageintent [\groupslot \textit{development}]]\\
$x_4$: \textit{can you send the preferred friends group} \\
$y_4$: [\sendmessageintent [\groupslot \textit{preferred friends} ] ] \\
$\hat{y}$: [\callintent [\groupslot \textit{Musicals}] ]\\
\end{tabular}
\caption{\label{table:paper_example_full} Input TF-IDF retrieves an exemplar with lexical overlap (`musicals') that is not relevant to the sample. The \comb retrieval balances lexical and label similarity and leads to a correct prediction.}
\end{table}

\end{document}